\documentclass[lettersize,journal]{IEEEtran}
\usepackage{amsmath,amsfonts}
\usepackage{algorithmic}
\usepackage{algorithm}
\usepackage{array}
\usepackage[caption=false,font=normalsize,labelfont=sf,textfont=sf]{subfig}
\usepackage{textcomp}
\usepackage{stfloats}
\usepackage{url}
\usepackage{verbatim}
\usepackage{graphicx}
\usepackage{cite}
\usepackage{cleveref}
\usepackage[export]{adjustbox}

\usepackage{listings}
\usepackage{xcolor}
\definecolor{changedcol}{rgb}{0.0,0.0,0.0}  
\DeclareCaptionFont{ccchngd}{\color{changedcol}}
\DeclareCaptionFont{ccblack}{\color{black}}
\newcommand{\Change}[1]{\color{changedcol}{#1}\color{black}}
\newenvironment{Changed}{%
	\captionsetup{textfont={ccchngd},labelfont={ccchngd}}%
\color{changedcol}%
}%

\lstdefinestyle{custom}{
    backgroundcolor=\color{gray!10},
    basicstyle=\ttfamily\footnotesize,
    frame=single,
    keywordstyle=\color{blue}\bfseries,
    commentstyle=\color{green!50!black},
    stringstyle=\color{red},
    numbers=left,
    numberstyle=\tiny\color{gray},
    stepnumber=1,
    numbersep=8pt,
    tabsize=4,
    showstringspaces=false,
    breaklines=true,
    breakatwhitespace=true
}

\begin{document}

\title{Towards Autonomous Reinforcement Learning \\for Real-World Robotic Manipulation \\with Large Language Models}

\author{Niccolò Turcato$^1$, Matteo Iovino$^2$, Aris Synodinos$^2$, Alberto Dalla Libera$^1$, Ruggero Carli$^1$ and Pietro Falco$^1$
\thanks{
This work has been submitted to the IEEE for possible publication. Copyright may be transferred without notice, after which this version may no longer be accessible.\\
$^1$ Department of Information Engineering, University of Padova, Italy.\\
$^2$ ABB Corporate Research, Västerås, Sweden.
}
}



\maketitle

\begin{abstract}
Recent advancements in Large Language Models (LLMs) and Visual Language Models (VLMs) have significantly impacted robotics, enabling high-level semantic motion planning applications. Reinforcement Learning (RL), a complementary paradigm, enables agents to autonomously optimize complex behaviors through interaction and reward signals. However, designing effective reward functions for RL remains challenging, especially in real-world tasks where sparse rewards are insufficient and dense rewards require elaborate design.
In this work, we propose \textit{Autonomous Reinforcement learning for Complex Human-Informed Environments} (\emph{ARCHIE}), an unsupervised pipeline leveraging GPT-4, a pre-trained LLM, to generate reward functions directly from natural language task descriptions. 
The rewards are used to train RL agents in simulated environments, where we formalize the reward generation process to enhance feasibility. Additionally, GPT-4 automates the coding of task success criteria, creating a fully automated, one-shot procedure for translating human-readable text into deployable robot skills. Our approach is validated through extensive simulated experiments on single-arm and bi-manual manipulation tasks using an ABB YuMi collaborative robot, highlighting its practicality and effectiveness. Tasks are demonstrated on the real robot setup.
\\
\end{abstract}


\section{Introduction}
\IEEEPARstart{T}{he} recent development of Large Language Models (LLMs), and Visual Language Models (VLMs) has drastically impacted academic and industrial research, due to their flexibility to many human-like tasks~\cite{hadi2023LLM_survey}. LLMs have found numerous applications in robotics~\cite{xiao2023robot_llm_survey}, including high-level semantic planning~\cite{saycan, singh2023progprompt, dalal2024plan, styrud2024automatic} and low-level locomotion and manipulation skills~\cite{pmlr-v229-yu23a_language2rewards, brohan2023rt, kwon2024llm_trajectory_gen}. 
The impact of LLMs and VLMs on robotic applications has opened new possibilities for how machines comprehend and interact with their environments. While these models demonstrate impressive high-level reasoning and language-based planning capabilities, the field of Reinforcement Learning (RL)~\cite{wang2022deepRL_survey} complements them by focusing on direct interaction with physical tasks. Due to its versatility and adaptability, RL has achieved astounding results in robotics~\cite{kober2013RL_robotics_survey} and other fields~\cite{zhang2021overview_rl_applications}, enabling agents to autonomously learn and optimize complex behaviors.

The main goal of RL algorithms is to generate and optimize autonomous agents to complete tasks, by autonomously interacting with the available environment. The agent receives feedback that evaluates its performance in the form of a numerical reward, and is trained to produce actions that maximize the expected rewards. 
RL approaches greatly reduce the effort required to manually define complicated tasks in environments with complex dynamics, often resulting in better performances than hand-coded approaches.
Nonetheless, careful design and engineering are required to define rewards to effectively and efficiently train RL agents.
Many real-world tasks can be encoded by sparse rewards, which are in general easy to define but result challenging to optimize with RL. Instead, in robotics and other fields dense rewards are highly adopted in robotics to guide the exploration of autonomous agents~\cite{Eschmann2021_reward_design} with the drawback of a more elaborate design. Indeed, dense reward functions can be difficult to tune, causing numerical instability, and misalignment w.r.t. the task objectives. Thus, automated or guided reward generation methods are interesting for both academic research and practical reasons.
Several recent publications have tried to handle the design of dense reward functions in an automated fashion by using state-of-the-art coding LLMs, such as GPT-4~\cite{achiam2023gpt4}. Their remarkable skills in code writing using natural language context allow non-expert users to define goals for RL agents as simple human-readable text. 
Such approaches reached remarkable performances in RL reward design in dexterous robotics manipulation~\cite{ma2023eureka, Nair2022, wang2023robogen, song2023self, xietext2reward}, locomotion~\cite{ryu2024curricullm,yao2024anybipe}, navigation~\cite{du2023guiding}, and autonomous driving~\cite{hazra2024revolve}.
Other applications of LLMs and VLMs for RL in robotics include approaches to guide the exploration process of RL agents~\cite{ma2024explorllm}, imitation~\cite{jin2024robotgpt, chen_rlingua_2024}, online fine-tuning~\cite{yang_robot_2024}, and Preference-based RL (PbRL)~\cite{wang2024prefclm}.
Remarkably,~\cite{szot2023large} employs a VLM as the backbone of a Deep policy, and RL training is used to fine-tune the policy performance for various tasks in simulated environments.
Current solutions for robotics manipulation, including~\cite{ma2023eureka,pmlr-v229-yu23a_language2rewards, wang2023robogen, song2023self, xietext2reward}, demonstrate few examples of applications in real-world settings, most requiring multiple training stages using feedback from human experts.
\begin{figure*}
    \centering
    \includegraphics[width=\textwidth]{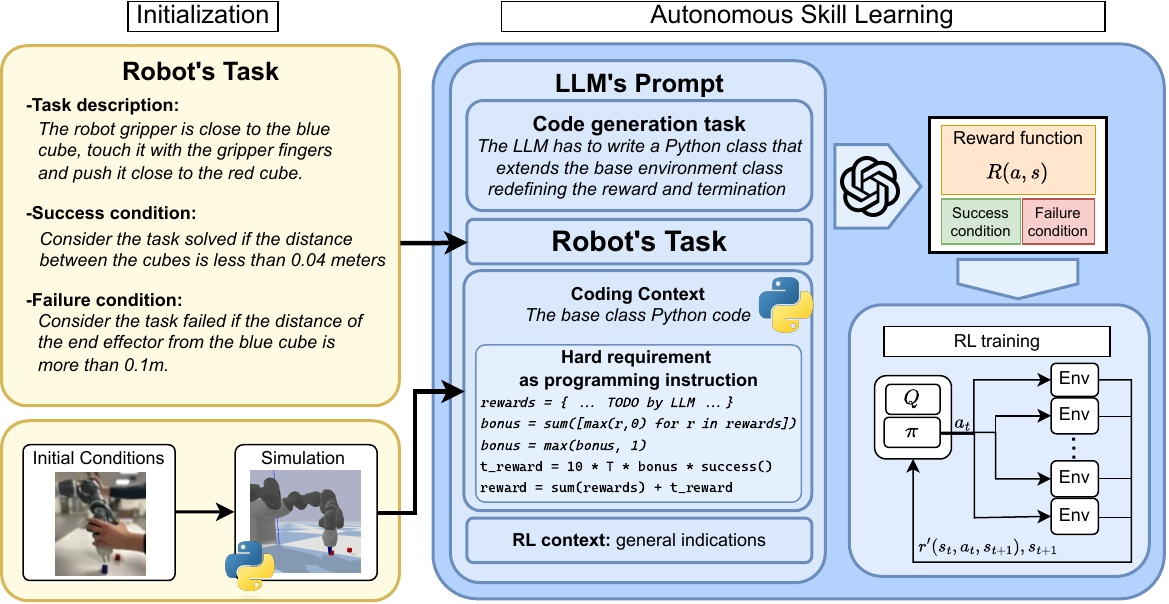}
    \caption{\Change{Overall schematic representation of \emph{ARCHIE},} composed of two phases: Initialization and Autonomous Skill Learning.}
    \label{fig:scheme}
\end{figure*}
In this work, we propose \textit{Autonomous Reinforcement learning for Complex Human-Informed Environments} (\emph{ARCHIE}). \emph{ARCHIE} is a practical automatic RL pipeline for training autonomous agents for robotics manipulation tasks, in an unsupervised manner. \emph{ARCHIE} employs GPT-4 \----a popular pre-trained LLM\---- for reward generation from human prompts.
We leverage natural language descriptions to generate reward functions via GPT-4, which are then used to train an RL agent in a simulated environment. Our approach introduces a formalization of the reward function that constrains the language model's code generation, enhancing the feasibility of task learning at the first attempt. 
Unlike previous methods, we also utilize the language model to define the success criteria for each task, further automating the learning pipeline. 
Moreover, by properly formalizing the reward functions in shaping and terminal terms, we avoid the need for reward reflection~\cite{ma2023eureka} and multiple stages of training in RL~\cite{song2023self}. \Change{This effectively reduces training time for RL agents, which is crucial, as training time impacts not only applicability of such methods, but also their energy and resource consumption.}
This results in a streamlined, one-shot process translating the user's text descriptions into deployable skills. 
\emph{ARCHIE} relies on the reasonable assumption that the manipulator's kinematic and dynamic models are available and that RL agents are trained in a simulated environment. 
We demonstrate the effectiveness of the reward generation in \Change{simplified simulated environments } and with a selection of real-world single-arm and bi-manual manipulation tasks on an ABB YuMi collaborative robot, shown in~\Cref{fig:gallery}.
Policies trained in simulation are demonstrated on the real robot in laboratory settings 
\footnote{videos available at \url{https://youtu.be/kaTxZ4oqD6Y}}, \Change{thus avoiding the use of safe RL methods.}.

This paper is organized as follows: \Cref{sec:related_works} discusses recent publications relevant to the scope of the paper, \Cref{sec:background} introduces the mathematical notations used in the rest of the paper. \Cref{sec:method} presents the proposed reward generation approach, \Cref{sec:experiments} shows the experimental results performed to validate the methods. \Cref{sec:conclusions} concludes the paper and discusses possible future integrations and lines of investigation.

\section{Related Works}
\label{sec:related_works}
In this section, we discuss relevant contributions with applications to robotic manipulation. 
We divide contributions into two main categories: (i) LLM-assisted RL and (ii) LLM for reward generation. The work presented in this paper falls into the second category.

\subsection{LLM-assisted RL}
In~\cite{ma2024explorllm}, the authors propose the use of a VLM and an LLM to guide policy learning in the context of robotics manipulation. The VLM is used as an open-vocabulary object detector to get the observation space of the RL policy, while the action space is defined w.r.t. the reference frame of the objects. The LLM is used to program both a high-level and a low-level policy to generate affordance maps~\cite{saycan} according to the visual input. The LLM policies are used during the training of a Soft Actor-Critic (SAC)~\cite{sac} agent with an $\varepsilon$-greedy exploration approach. 
In~\cite{jin2024robotgpt, chen_rlingua_2024}, the authors use the LLMs to define expert agents to guide the RL training with imitation.  The approach in~\cite{jin2024robotgpt} relies on the availability of \emph{pick}/\emph{place} primitives that make the RL agent's action space discrete. The LLM policies generate expert demonstration data that bootstrap the training of an SDQfD~\cite{SDQfD} agent.
In~\cite{chen_rlingua_2024}, a variation of Twin Delayed Deep Deterministic Policy Gradient (TD3)~\cite{td3} is proposed, to accommodate an imitation loss in the actor's loss. The imitation data is collected during the training process with a robot controller programmed with an LLM.
In~\cite{yang_robot_2024} the researchers used Cal-QL~\cite{cal-ql}, a two-step policy training composed of (i) an offline stage where a policy is pre-trained to imitate various demonstrations, (ii) an online stage where the policy is fine-tuned to solve specific tasks while using a VLM to generate rewards directly from visual observations. 
In~\cite{wang2024prefclm} authors incorporate PEBBLE~\cite{lee2021pebble} with natural language user feedback on the iteratively learned robot policies during the PbRL training. Preferences of multiple LLMs are collected to compute a scoring of the performance of agents.

\subsection{LLM for reward generation}
Most notably, Eureka~\cite{ma2023eureka} is one of the most recent contributions in reward generation for RL. It employs GPT-4 to generate dense reward functions in the form of Python code by using the environment's source code and a natural-language task description as context. Then, it iterates the training and reward reflection by querying the LLM with feedback from the previous episode to improve the performance of the agents. Such a strategy has proven to be effective in simulation, where it can accomplish various robotic tasks that cannot be solved in a sparse reward setting. Remarkably, the reward functions generated by GPT-4 prove to be more robust than rewards generated by humans. Recently, Eureka was applied in a real-world robotic locomotion task~\cite{ma2024dreureka}, but to the best of our knowledge, not to robotic manipulation.
Similarly, in~\cite{song2023self} and~\cite{xietext2reward}, the authors use an LLM to design reward functions based on natural language descriptions of the task and environment. \Change{The initial reward function design is then used to train an RL agent, and the results are fed back to the LLM for self-refinement. } This iterative process continues until the reward function meets the desired performance criteria. Notably,~\cite{xietext2reward} presented results in real-world dexterous manipulation tasks.
In~\cite{Nair2022}, authors use Offline RL~\cite{prudencio2023survey_offlineRL} to train language-conditioned robot behaviors from large annotated offline datasets that outperform language-conditioned imitation learning techniques. The method exhibits zero-shot generalization to unseen natural language commands in real-world settings.
The work in~\cite{wang2023robogen} leverages LLMs and VLMs to extract knowledge about object semantics and common-sense task relevance to autonomously and continuously generate new skills through a self-guided learning cycle. In this cycle, the VLM-powered agent autonomously decides the learning task and prepares a natural language description followed by a cost or reward function. Remarkably, it also generates an appropriate simulated environment using a dictionary of available objects. Tasks are solved with RL, Model Predictive Control (MPC)~\cite{schwenzer2021review_MPC} or other Optimal Control strategies. With this approach, the agent builds a plethora of skills for a variety of self-proposed tasks.

\vspace{.1cm}
While showing promising results, including experiments using real-world manipulators, these works require multiple training stages or annotated demonstrations. For these reasons, we propose a pipeline with a single training stage, making it more practical for real-world applications.

\section{Notations}
\label{sec:background}
Reinforcement Learning (RL) problems can be described as Markov Decision Processes (MDPs), defined by the tuple $(\mathcal{S}, \mathcal{A}, \mathcal{P}, R, \gamma)$. 
$\mathcal{S}$ and $\mathcal{A}$ are, respectively, the state space and the action space. In case both $\mathcal{S}$ and $\mathcal{A}$ are continuous, the transition density function $\mathcal{P}$ is continuous and formalized as $\mathcal{P}: \mathcal{S} \times \mathcal{A} \times \mathcal{S} \rightarrow [0, \infty)$. $r: \mathcal{S} \times \mathcal{A} \times \mathcal{S} \rightarrow \mathbb{R}$ is the random variable describing the reward function, and $\gamma \in [0,1]$ is a discount factor.
At each time-step $t$, the policy function $\pi$ maps the current state $s_t \in \mathcal{S}$ to an action $a_t \in \mathcal{A}$ with respect to the conditional distribution $\pi(a_t|s_t)$. 
\newline
RL algorithms aim at optimizing the policy $\pi$ to maximizes the expected discounted sum of rewards $R_0 = \mathbb{E_{\pi_\phi}}\left[\sum_{t=0}^\infty \gamma^t r(s_t, a_t, s_{t+1})\right]$ of a specific MDP. \Change{The main mathematical tool designed to solve RL tasks is action-value function $Q$:}
\begin{align}
Q^\pi(s_t, a_t) &= \mathbb{E_{\pi}}\left[R_t | S_t = s_t, A_t = a_t\right]
\end{align}
Then, the final objective can be defined as finding an optimal policy $\pi^*$ such that
\begin{equation}
    \pi^* = \text{argmax}_{\pi} \left\{ \mathbb{E}_{s_0 \sim p(s_0)} \left[ Q^{\pi}(s_0, \pi(s_0)) \right] \right\},
\end{equation}
given an initial state distribution $p(s_0)$.

\section{Method}
\label{sec:method}
\emph{ARCHIE}, depicted in~\Cref{fig:scheme}, comprises two main stages. In the first stage, namely the initialization stage, described in~\Cref{sec:simgen,sec:nl2r}, we set the initial condition of the robot and environment in the simulation, to match the real world. Moreover, we collect the user text input composed of the task description, failure, and success conditions. The second stage, named the Autonomous Skill Learning stage, prompts GPT-4 with the user description to generate code defining reward function and termination conditions, according to the formalization described in~\Cref{sec:dense_reward_form}. The generated reward is used to train an RL agent, as presented in~\Cref{sec:policy_learn}.

\subsection{Initialization: Simulation Generation}
\label{sec:simgen}
For most manipulation tasks, simulation generation can be done manually by hand-placing 3D models in a simulation environment. Nonetheless, automatic simulation generation, as proposed in~\cite{wang2023robogen}, would allow further automation of a skill-learning process, thus requiring much less human intervention. To this end, we employ a methodology for automatic simulation generation composed of two Deep Learning models: (i) YoloWorld~\cite{cheng2024yolo} for object detection and (ii) NanoSAM~\cite{nanosam} to segment the objects of interest from a 3D camera image and depth plane. This allows us to extract a point cloud representation of the object which is then used to compute the coordinates w.r.t. the robot frame and populate the simulation with an available 3D model. The YoloWorld model requires a text descriptor of the objects to detect (e.g., ``box", ``cube") which the LLM can infer from the user's task description.

\subsection{Initialization: Natural Language to Rewards}
\label{sec:nl2r}
Similarly to previous work, \emph{ARCHIE} exploits the context-aware code-writing capabilities of modern LLMs for reward generation in the context of robot manipulation tasks. 
To this end, we require a base Python class that implements the basic functionalities to simulate a manipulation task, namely:
\begin{itemize}
    \item Simulation engine and robot control law.
    \item Application Programming Interfaces (APIs) for contact and collision checks.
    \item APIs for reading objects and end effectors poses.
\end{itemize}
Given a task for the robot, the LLM is used to write a new Python class that extends the base class by defining the dense reward function to guide the agent's exploration towards the task goal.
Unlike previous publications, we require the LLM to implement the termination conditions for the defined task, including a success condition that evaluates whether the task is solved and a failure condition that assesses whether the agent has failed. 
This is done to reduce the ambiguity in task definitions in natural language, as the geometrical context is very complex to infer from the environment's code. 
Moreover, the failure condition reduces the amount of exploration required by the agents, since failing states are associated with low rewards and are not interesting to explore. In this way, the agent is encouraged to avoid failure instead.
\Change{In \Cref{fig:scheme} (yellow block) we report an example of user task description, including clear geometrical definitions of success and failure for a cube pushing task. }
\Change{With clear geometrical indications, writing the termination conditions becomes a trivial programming task, which the LLM can solve. }

The natural language prompt is composed of the following blocks:
\begin{itemize}
    \item Introduction, in which the overall LLM's \Change{code generation } task is described, \Change{including the expected format of the output Python code};
    \item Robot's task description, including a high-level description and success and fail conditions with details about the geometry of the task;
    \item \Change{Coding Context}, including the Python code and documentation of the base \Change{environment class, and the instructions to implement the dense reward formalization in \Cref{sec:dense_reward_form}}. The documentation is an important part of the context as it explains how to use the APIs to register contacts and poses of the objects useful for defining the reward components.
    \item \Change{RL Context: } general indications on reward functions for manipulation tasks, including some examples.
\end{itemize}
\Change{The prompt design is overviewed in \Cref{fig:scheme} (blue box), and the full text is reported in the PDF appendix. }
\Change{As in previous work, the code generation instructions are generic and high level, as examples of RL code, including rewards, are present in modern LLM training data.}

\subsection{Autonomous Skill Learning: Dense reward formalization}
\label{sec:dense_reward_form}
To guide agents' exploration, the reward function is divided into two main terms: (i) the shaping term and (ii) the terminal reward. We refer to a generic RL setting where the agent performs a series of attempts at solving the task, also called episodes, until a maximum of total timesteps is reached. We reasonably assume that the episodes have a fixed maximum length of $T \in \mathbb{N}$ timesteps.

Let $r^k(s_t,a_t)$, $k=1, \dots, K$ be the $k$-th reward component of our dense reward function shaping, received after applying action $a_t$ in state $s_t$. 
These terms are generated by the LLM using the prompt engineering defined in~\Cref{sec:nl2r} \Change{and overviewed in \Cref{fig:scheme}}.
Arbitrarily, the reward components $r^{k_+}(s_t,a_t)$, $k_+ \in \boldsymbol{K_+} \subseteq \{1, .., K\}$ are positive and are called bonuses.
Due to the available context, the shaping terms are generally coherent with the task requirements. Nevertheless, there is no guarantee that the different generated terms are numerically coherent with each other, as discussed in other works. 
In fact, the authors in~\cite{ma2023eureka, song2023self} propose iterative processes in which the LLM evaluates the results of the trained agents and adjusts the weights of the different addends. 
In an effort to avoid this iterative process, we introduce a clearly defined terminal reward term $R_F(s_t, a_t)$ and two binary classifiers $\Phi(s_t)$ and $\Gamma(s_t)$, \Change{which implement respectively the success and failure conditions checks on state $s_t$. }

\begin{equation}
\label{eq:step_reward}
    r'(s_t, a_t, s_{t+1}) = \sum_{k=1}^K r^k(s_t, a_t) + R_F(s_t, a_t) \cdot \Phi(s_{t+1})
\end{equation}

The terminal reward is defined as  
\begin{equation}
\label{eq:terminal_reward}
    R_F(s_t, a_t) = 10 \cdot T \cdot \text{max}(\sum_{k \in \boldsymbol{K_+}}r^{k_+}(s, a), 1)
\end{equation}
To ensure that the terminal reward is sufficiently high in any terminal state $s_{t'+1}$, the sufficient condition is
\begin{equation}
\label{eq:condition_reward}
    r^{k_+}(s_{t'}, a_{t'}) \geq r^{k_+}(s_{t}, a_{t},) \quad \forall t=0,\dots,t', {k_+} \in \boldsymbol{K_+}
\end{equation}
which is reasonable for reward shaping.

\textbf{Statement}: If \cref{eq:condition_reward} holds, then the terminal reward $R_F(s_{t'}, a_{t'})$ for a success state $s_{t'+1}$ (i.e. $\Phi(s_{t'+1})=1$) is at least one order of magnitude greater than the sum of rewards collected in the trajectory $(s_0, \dots, s_{t'})$, i.e. $R_F(s_{t'}, a_{t'}) >>  \sum_{t=0}^{t'}r(s_t, a_t)$

\textbf{Proof of statement}:\\
The statement \Change{is true } if $\sum_{k_+ \in \boldsymbol{K_+}} r^k(s_{t'}, a_{t'}) \leq 1$, otherwise it is sufficient to apply \cref{eq:condition_reward} to the definition in \cref{eq:terminal_reward}:
\begin{align}
    10  T  \sum_{k_+ \in \boldsymbol{K_+}} r^{k_+}(s_{t'}, a_{t'}) \cdot  &>> \sum_{t=0}^{t'} \sum_{k_+ \in \boldsymbol{K_+}} r^{k_+}(s_t, a_t) \\
    \sum_{t=0}^{t'} \sum_{k_+ \in \boldsymbol{K_+}} r^{k_+}(s_t, a_t) &> 
    \sum_{t=0}^{t'}r(s_t, a_t) \quad \ensuremath{\Box}
\end{align}

In this way, we guarantee that the terminal reward for completing the task is always at least one order of magnitude greater than the cumulative rewards that the agent can collect during the trajectory before termination. For $\gamma < 1$, this encourages the agent to move toward the terminal states rather than stagnating in high-rewarding non-terminal states, which is likely to happen if the shaping components are not balanced.
\Change{The formalization is enforced as a hard programming constraint in the LLM's prompt, as reported in \Cref{fig:scheme} (blue box), and in the PDF appendix.}\\
\Change{\textbf{Remark:} this hard constraint on the code generation is the main difference w.r.t. previous work, e.g. Eureka~\cite{ma2023eureka}.}

 \begin{figure}
    \centering
    \subfloat{\includegraphics[width=\columnwidth]{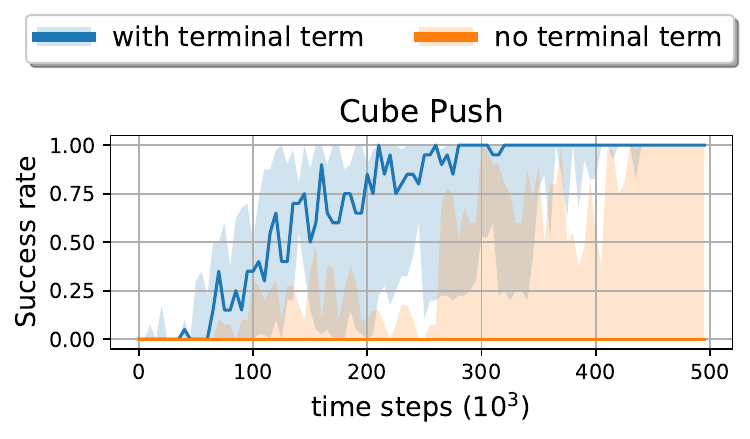}}%
    \caption{\Change{Cube push example trainings. SAC trained with rewards generated by GPT-4, with and without the proposed formalization.}}
    \label{fig:pushing_example}
\end{figure}
\subsection{Autonomous Skill Learning: Policy Learning}
\label{sec:policy_learn}
The reward function formalized in~\Cref{sec:dense_reward_form} is used to train an RL agent in simulation. We use a slightly modified version of SAC~\cite{sac}, where the agent explores in parallel $N$ simulated environments and the actor updates are delayed from the critic updates, as in TD3~\cite{td3}. Both modifications are very commonly employed to speed up the learning process in terms of wall clock time. Parallelization allows the agent to explore more of the state space in the same time as the standard version of the algorithm. The delayed actor update lowers the overall computational burden of the algorithm.

The pseudocode of the policy learning algorithm is reported in \Change{the PDF appendix (\Cref{alg:sac_delays})}, it follows the notations used in~\cite{sac, td3} and related research. \Change{Success and failure conditions $\Phi$ and $\Gamma$ are used to terminate episodes. }
In the simulated environment, the agent's state is defined as a vector containing the pose and linear and angular velocities of the robot's end-effectors, the position and velocity of the grippers' joint, and the pose of the objects in the scene. The agent's action is defined as the reference linear and angular velocities of the end-effectors and the reference velocity of the grippers' joint. This definition of state and action spaces allows the agent to directly correlate actions with changes in the state and reward. For single-arm tasks, the state and action spaces are limited to drive the left arm, while the right arm is fixed to a constant joint configuration.

\begin{figure*}
    \centering
    \subfloat{\includegraphics[width=\linewidth,valign=t]{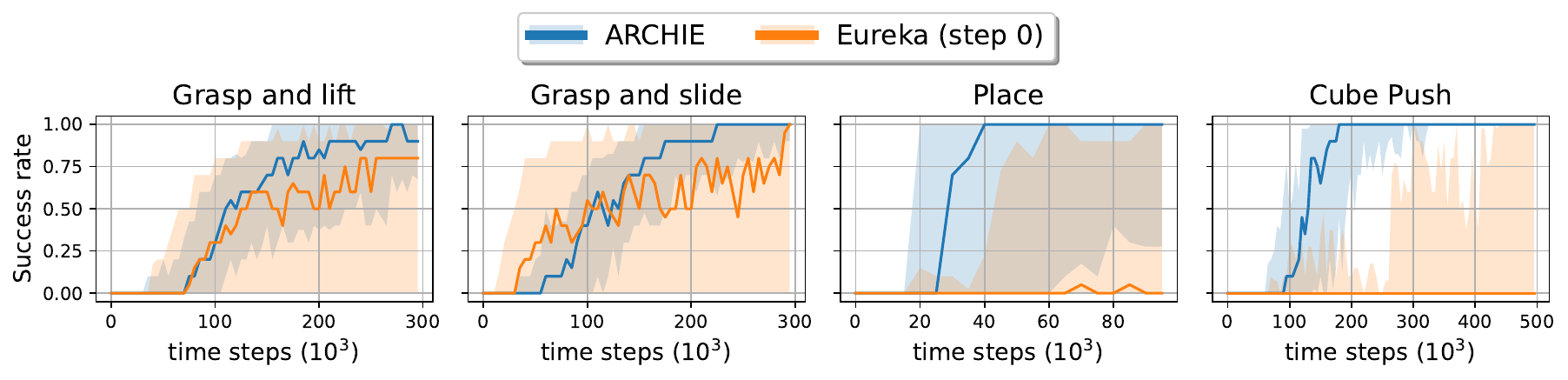}}%

    \subfloat{\includegraphics[width=\linewidth,valign=t]{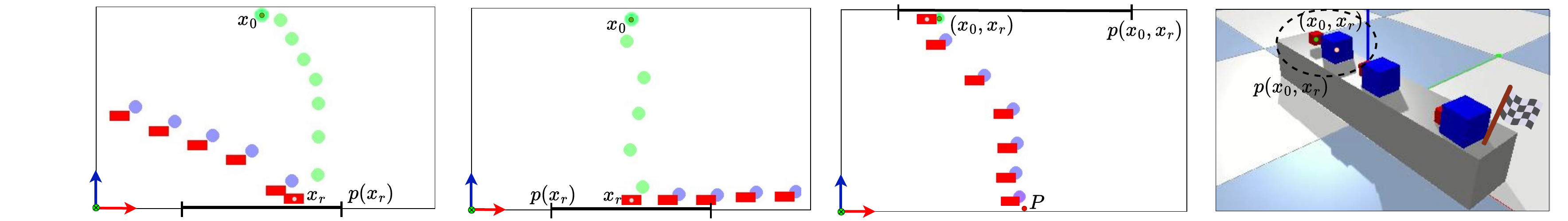}}%
    \caption{Top: Success rates for SAC agent training on the selected benchmark tasks. The proposed reward formalization is compared with the performance of the rewards completely defined by \Change{Eureka at step 0 with } GPT-4. For each task, 10 rewards are generated and each reward is tested with 10 different random seeds. Plots show median and 25-75 percentiles. \Change{Bottom: The agents performing the tasks. In the 2D environment, the agent (ball) is purple when holding the object (rectangle), while green otherwise.}}
    \label{fig:2d_envs}
\end{figure*}

\section{Experiments}
\label{sec:experiments}
In this section, we present the experimental evaluations performed to validate the proposed method. First, we validate the effectiveness of the reward formalization described in \Cref{sec:dense_reward_form} on three tasks on a simple 2D and 3D simulation. Second, we present the results of the Autonomous RL pipeline for 10 different robotics manipulation tasks, including demonstrations on the real setup.

In all experiments, the SAC agents are trained using Networks of $512$ neurons and $3$ layers, with \emph{ReLU} activation functions, for both policy and action-value function. In training, during exploration, actions are sampled from the stochastic policy's posterior distribution. In evaluation, the actions are computed as the mean of the policy's distribution. SAC's entropy parameter is tuned online as in \cite{haarnoja2018sac_entropy_tuning}. Each evaluation is obtained averaging over 10 episodes, varying the random seed. The discount factor is set to $\gamma=0.99$ and the soft update coefficient is set to $\tau=5\cdot10^{-3}$. The duration of the episodes is set to $T=10^3$, simulating with a timestep $dt=10^{-2}$s.

\subsection{\Change{Benchmark } Environments }
\label{sec:bench_experiments}
To assess the effectiveness of the proposed reward formalization, we implemented \Change{simplified simulations } that mimic the simplified dynamics of a robotics manipulation scene. \Change{In a first environment}, the agent controls the movements of a point on a 2D vertical plane, a rectangular object is present on the ``floor'' of the plane. When in proximity to the object, the agent can perform grasping, which constrains the object to move solidly with the agent, until it stops the grasping, releasing the object. 
We consider three simple manipulation tasks: (i) a grasp and lift task, (ii) a grasp and slide task, (iii) a placing task. \Cref{fig:2d_envs} (bottom) shows an example execution for each task in the 2d environment. 
In the first two tasks, the agent is initialized to a constant position $x_0$, while the object is initialized in a position $x_r \sim p(x_r)$. In the third task, the object and the agent are initialized next to each other, randomly at the top of the environment, $(x_0, x_r) \sim p(x_0, x_r)$. \\ 
\Change{In a second environment, implemented with 3D bullet simulation engine, the agent controls the movements of a small box, and can interact with a bigger box, over a narrow eleveted plane, with pushing actions. In this environment we consider a pushing task with constraint: the agent is required to push the object from one side of the surface to the other, without pushing the object from the surface. \Cref{fig:2d_envs} (bottom) shows an example of task execution in the 3D simulation. \\}
\Change{For each task, we train agents with \emph{ARCHIE} and a baseline reward generation. The selected baseline is the first step of reward generation of Eureka \cite{ma2023eureka} and similar works \cite{song2023self}, which is equivalent to the reward generation of \emph{ARCHIE}, minus the formalization defined in \Cref{sec:dense_reward_form}. This comparison serves to highlight the improved learning stability due to the reward formalization. \\} 
We generate 10 rewards for both reward generations, using the same prompt instructions, except for the part that defines the formalization of the reward. 
\Change{The prompt structure and the natural language task definitions are available in the PDF appendix. }
We train 10 SAC agents for each reward using different random seeds and monitor the success rates of the trained policies. We use a single training environment ($N=1$) per agent. The results are reported in~\Cref{fig:2d_envs} (top), where we show the statistics of the success rate curves of the agents trained by \emph{ARCHIE} and the statistics of the rewards generated by \Change{Eureka (step 0)}.
\\
In the training horizons considered, the agents trained by \emph{ARCHIE} with the proposed formalization consistently solve all tasks, while the agents trained with \Change{Eureka (step 0) } are much less consistent.
This suggests that the proposed formalization effectively stabilizes policy learning toward task completion.\\
\begin{Changed}
\textbf{\emph{Running example:}} We remark one example of reward generated by \emph{ARCHIE} for the Cube Push task in the 3D environment (\Cref{fig:2d_envs} bottom far right):
\begin{equation}
    r(s_t, a_t, s_{t+1}) = -d + 10 \cdot contact + x_t + \underbrace{R_F(s_t, a_t) \Phi(s_{t+1})}_{\text{terminal term}},
    \label{eq:reward_example}
\end{equation}
Here $\Phi(s_{t}) = \text{step} (x_t - table\_lenght)$, $d$ is the distance between the two cubes, $contact$ is an observed variable which is true when the cubes are in contact, and $x_t$ is the x coordinate of the blue cube. This is a typical example of reward with correct structure and unbalanced weights, since $d$ and $x_t$ are both in the order of the unit, while the weight of $contact$ is one order of magnitude greater. The Python code that implements this reward can be found in the PDF appendix. In \Cref{fig:pushing_example} we show the learning curves of SAC agents under the reward in \Cref{eq:reward_example}, as well as the same reward without the terminal term. Indeed, the agents trained with the terminal reward all converge to successful solutions, instead the other agents fail. Again, this further highlights the benefits of the proposed formulation.
\end{Changed}

\subsection{Robotic Manipulation}
We select 10 robotic manipulation tasks, \Change{ each identified by a codename, the detailed task descriptions used for the LLM's prompts are included in the PDF appendix. Single arm tasks: (i) CubeSlide: push a blue cube close to a red cube; (ii) CubePick: pick the blue cube without hitting the red cube next to the first cube; (iii) CubeStack: the robot is holding a blue cube, stack the blue cube over the red cube; (iv) CubesTowerSlide: push two stacked cubes 10cm to the right; (v) VialGrasp: Grasp a vial from a vial holder; (vi) CentrifugeInsertion: the robot is holding the vial, insert the vial in a centrifuge. Dual arm tasks: (i) BimanualPenInTape: one gripper is holding a pen and the other is holding a tape, insert the pen in the tape's hole; (ii) CubeInCup: one gripper is holding a cube and the other is holding a cup, place the cube into the cup; (iii) BimanualBoxLift: Lift a box with both grippers; (iv) BimanualHandover: one gripper is holding a box, the other is empty, hand over the box to the empty gripper. }
For each task, we define the natural language task description, as described in \Cref{sec:nl2r}, which we use to generate a reward with \emph{ARCHIE}'s proposed reward generation. The same descriptions are exploited to generate code policies \cite{liang2023code} with GPT-4, using the same interface as the RL policies.
\begin{figure*}[h]
    \centering
    \subfloat{
    \includegraphics[width=\linewidth]{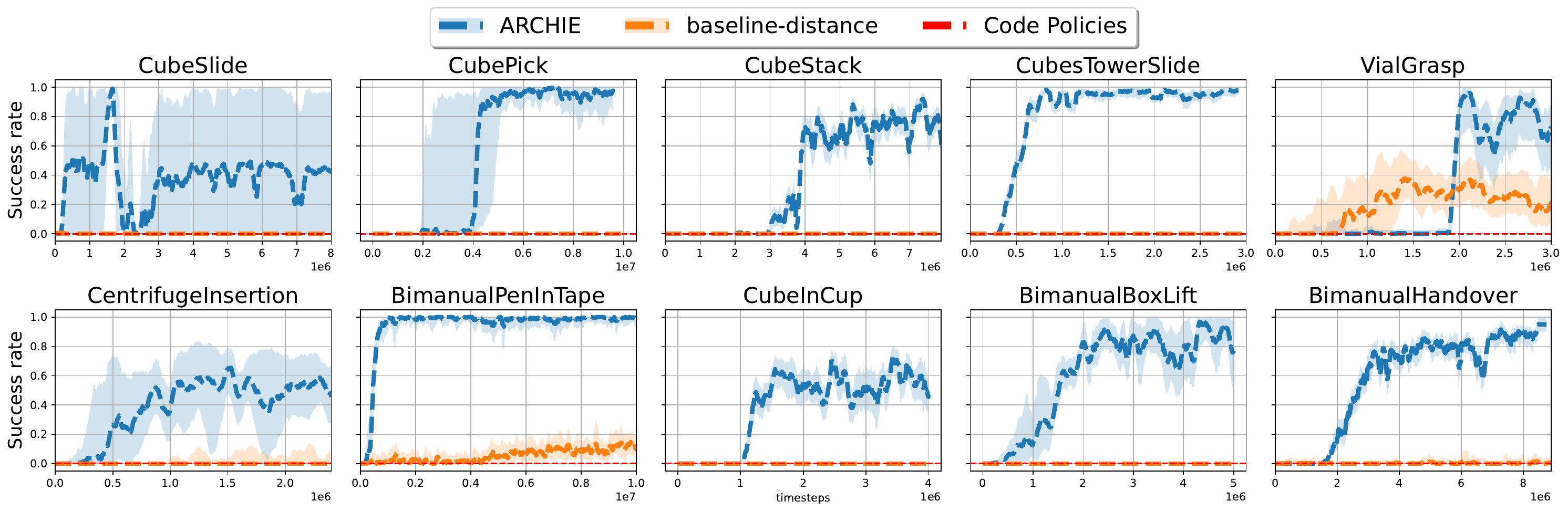}}%
    \caption{Success rate curves of the SAC agents during the training procedure. Curves and shaded areas show the median and 25-75 percentiles of the success rates. Each agent is evaluated on each parallel environment, with 10 policy execution, every 5K steps of training.}
    \label{fig:rl_curves}
\end{figure*}
\begin{figure}[h]
    \centering
\includegraphics[width=\columnwidth]{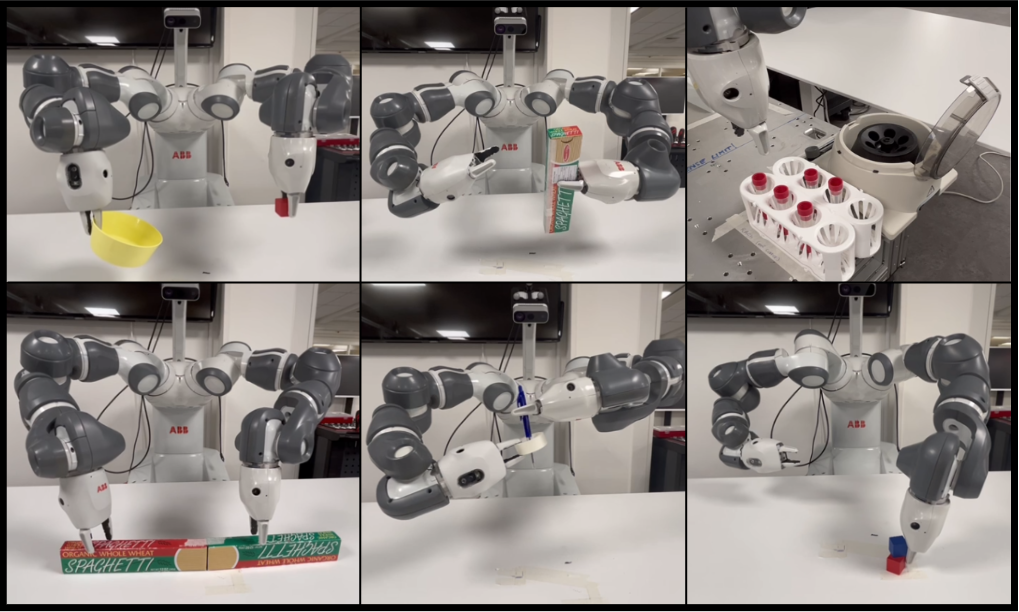}
    \caption{The YuMi robot performing some of the selected tasks.}
    \label{fig:gallery}
\end{figure}
\label{sec:rob_manipulation}
As baseline for comparison, for all tasks, we manually define a purely distance-based reward function, inspired by~\cite{9211756,levine2015learning}. 
In this case, we do not compare with~\cite{ma2023eureka, song2023self} or similar approaches that require multiple training steps, since the proposed approach is a one-shot procedure.
\\
Each reward function, including the baseline, is used to train 3 SAC agents, for each task, using different random seeds. Each agent explores $N=4$ environments in parallel, as described in \Cref{sec:policy_learn}. 
\\
The training results are reported in \Cref{fig:rl_curves}, where we compare the agents in terms of success rate.
The figure includes the average success rates of 10 different code policies for each task. The code policies do not solve any of the selected tasks, while the rewards generated by \emph{ARCHIE} guide the agents toward task completion in all tasks, reaching high success rates in most tasks. \Change{In some tasks, the RL training under \emph{ARCHIE} presents some instability, which could be due to the discontinuity in the reward function. }
Instead, the distance-based baseline rewards only solve a few of the tasks, with low success rates.
These results suggest that the reasoning capabilities of the LLM are adequate to select the reward terms related to the objective task, while it would practically struggle to define the low-level skills required to solve these tasks without ad-hoc interfaces. 
Moreover, a proper mathematical structure in the \Change{formalization of the rewards} is fundamental to exploit the code generation capabilities of the LLM. 
\\
\Cref{fig:gallery} depicts the YuMi robot performing some of the tasks in the lab settings, \Change{while the experiments are shown } in the demonstration video. The robot is controlled with a joint position controller, tracking the trajectories performed by the policy trained in simulation. The gripper is operated with a combination of force/position control.

\section{Discussion and Conclusions}
\label{sec:conclusions}
In this work, we proposed \emph{ARCHIE}, an unsupervised pipeline leveraging GPT-4 to automate reward function generation for Reinforcement Learning (RL) in robotic manipulation. By utilizing Large Language Models (LLMs), our method translates human-readable task descriptions into structured reward functions, enabling training of RL agents in simulated environments with minimal manual intervention. Unlike previous methods, our approach integrates both reward generation and task success criteria formulation, creating a one-shot process for autonomous skill acquisition\Change{, significantly reducing the required training time.}

Through extensive experiments on simulated robotic manipulation tasks using an ABB YuMi robot, we demonstrated the effectiveness of our method in solving various complex tasks. 
Our results showed that policies trained using LLM-generated rewards achieved comparable or superior performance to well-known human-engineered rewards, such as the ones in~\cite{9211756,levine2015learning}. Meanwhile, the time and expertise required for reward design are significantly reduced.
In the training horizons considered, the agents trained by \emph{ARCHIE}, with the proposed formalization, consistently solve all robotics tasks, while the agents trained with the human-engineered rewards are much less consistent. All considered tasks are not solved by code policies generated by \emph{GPT-4}.

Furthermore, our proposed reward formalization ensures stable policy learning, at the first training attempt, by balancing shaping and terminal rewards, mitigating issues related to reward misalignment.
In contrast, other works rely on reward reflection to iteratively improve the generated rewards, thus requiring several RL policy training attempts and prompts to the LLM to converge to a stable policy.
In the \Change{benchmark } environments, we showed that the proposed one-shot reward generation is much more effective than the first step of reward generation \Change{of Eureka~\cite{ma2023eureka}, completely formalized by \emph{GPT-4}, as also done in other works, such as \cite{song2023self}}.

While our approach significantly advances automated RL training, several avenues remain unexplored for future research. The introduction of reward function refinement through real-world feedback and human preferences may allow \Change{for the integration of the } training pipeline in robot programming software.
\Change{The integration of safe RL strategies in real-world fine-tuning could improve sim-to-real transfer for realistic applications. }
Finally, exploring more complex \Change{robotic } scenarios and multi-agent learning setups could extend the applicability of our method to a broader range of real-world tasks.
Our findings highlight the potential of LLMs in accelerating RL-based robotic learning, paving the way for more autonomous and efficient robot training paradigms.




\bibliographystyle{IEEEtran}
\bibliography{ref}

\newpage

 
\vspace{11pt}

\vfill

\clearpage
\onecolumn

\section*{Appendix}
\label{sec:appendix}

\subsection{Pseudocode}

\begin{algorithm}
\caption{Parallel SAC with delayed actor updates}
\label{alg:sac_delays}
\begin{algorithmic}[5]
   \STATE Initialize critic $Q_{\theta_1}, Q_{\theta_2}$, and actor $\pi_\phi$ networks
   \STATE Initialize target networks $\theta_1' \leftarrow \theta_1, \theta_2' \leftarrow \theta_2, \phi' \leftarrow \phi$
   \STATE Initialize replay memory \textbf{$\mathcal{B}$} 

    \FORALL{$i \in \{1, \ldots, N\}$ \textbf{in parallel}}
        \STATE $j=0$
        \REPEAT
            \STATE $t=0$, $s_0 \sim p(s_0)$, $done=$False
            \REPEAT
                \STATE Sample action $a_t \sim \pi_{\phi'}(s_t)$, $s_{t+1} = \mathcal{P}(s_t,a_t)$.
                \STATE $r = r(s_t,a_t)$ (\cref{eq:step_reward})
                \STATE $done=\Phi(s_{t+1})$  or $\Gamma(s_{t+1})$ or $t+1 == T$

                \STATE Store $(s_t,a_t,r,s_{t+1},done)$ tuple in $\mathcal{B}$
            \STATE Update critics $(\theta_1, \theta_2)$
            \STATE $\theta_i' \leftarrow \tau \theta_i + (1-\tau)\theta_i'$, $\forall i=1,2$
            \IF{$j$ mod $d$}
                \STATE Update actor $\phi$ 
                \STATE $\phi' \leftarrow \tau \phi + (1-\tau)\phi'$
            \ENDIF
            \UNTIL{not $done$ and $j + t < T_{max}$}
            \STATE $j \leftarrow j + t$
        \UNTIL{$j < T_{max}$}
        

    \ENDFOR

\end{algorithmic}
\end{algorithm}
\newpage

\subsection{GPT-4 prompts for reward generation}
The task required for the GPT-4 model is a code generation task, in this appendix, we report the prompt structure we used. 
\noindent\fbox{%
    \parbox{\textwidth}{%
    You are a reward engineer trying to write reward functions to solve reinforcement learning tasks as effective as possible. Your goal is to write a dense reward function for the environment that will help the agent learn the task described in text. Your reward function should use useful variables from the environment.
    
    \vspace{5pt}
    
    Write a python class that satisfies the following requirements:

    -The new class must inherit from the class defined in the code snipped included after the requirements.
    
    -The new class must not redefine the constructor
    
    -The new class must only redefine the reward\_fun and \_get\_info methods to encode the following task for the environment:
    
    \begin{center}
        TASK DESCRIPTION
    \end{center}

    -The reward\_fun method must define a dictionary named reward\_dict, reward\_dict must contain all the values of the reward components, excluding the reward for solving the task. 
    
    \vspace{5pt}
    
    Then the method must conclude the following code:

    {\color{blue}
    {\footnotesize \texttt{total\_bonuses = sum([rewards\_dict[k] if rewards\_dict[k] $>$ 0 else 0 for k in rewards\_dict.keys()])}}
    
    {\footnotesize \texttt{total\_bonuses = max(total\_bonuses, 1)}}
    
    {\footnotesize \texttt{task\_solved\_reward = 10 * self.\_max\_episode\_steps * total\_bonuses * info['task\_solved']}}
    
    {\footnotesize \texttt{total\_shaping = sum([rewards\_dict[k] for k in rewards\_dict.keys()])}}}
    
    {\footnotesize \texttt{reward = total\_shaping {\color{blue}+ task\_solved\_reward}}}
    
    {\footnotesize {\color{blue} \texttt{rewards\_dict['task\_solved\_reward'] = task\_solved\_reward}}}
    
    {\footnotesize \texttt{return reward, rewards\_dict}}
    
    \vspace{5pt}
    
    The \_get\_info method must return a dict that contains a boolean field task\_solved, that is computed according to the task description provided.
    If the prompt requires, the new class can implement a termination\_condition method, that returns True if a failure condition is detected.
    Not all tasks require this failure condition to be implemented. 
    \begin{center}
        \texttt{BASE CLASS CODE}
    \end{center}

    \vspace{5pt}
    Here is some advice on how to design reward functions: 
    
    -The reward function should implement a shaping that clearly guides the policy towards the goal
    
    -You can give a rewarding bonus if the task is partially solved.
    
    -If the environment has a termination condition that can halt an episode before reaching the goal area or configuration, consider using positive shaping terms in the reward function.
    
    -The reward components should not be too disproportionate.
    
    -To incentive to get close to an object you should reward the decrease of distance and the contact with said objects, if you want to avoid touching another object just give negative rewards if that is touched.
    
    -If you want to grasp an object and/or not to drop it you should reward contacts with the gripper fingers and the object
    
    -To reward lifting of objects, you can assume that the 3rd dimension of position vectors is the vertical axis, which is oriented towards the ceiling. You can use +height as reward shaping for lifting tasks.
    
    
    -Penalties for unwanted actions should be very small in absolute value, compared to positive rewards.
    
    \vspace{5pt}
    
    Here is one example of reward function for a cube picking task:
    \vspace{5pt}
    
    {\footnotesize \texttt{cube\_pose = np.array(sim\_utils.getObjPose(self.cube\_id))}}
    
    {\footnotesize \texttt{end\_effector\_left\_pose = np.array(sim\_utils.getLinkPose(self.robot\_id, self.end\_effector\_left\_idx))}}
    
    {\footnotesize \texttt{distance = np.linalg.norm(np.array(cube\_pose[:3]) - np.array(end\_effector\_left\_pose[:3]))}}
    
    {\footnotesize \texttt{finger\_1, finger\_2, total\_contact\_force = self.check\_fingers\_touching(self.cube\_id)}}

    {\footnotesize \texttt{rewards\_dict = \{ }}
    
    {\footnotesize \texttt{"distance\_to\_cube\_reward": -distance,  \# encourage getting close to the target cube }}
    
    {\footnotesize \texttt{"finger\_contacts\_reward":  1 * finger\_1 + 1 * finger\_2,  \# encourage interaction with the target cube }}
    
    {\footnotesize \texttt{"cube\_height\_reward": 1000 * (cube\_pose[2] - 0.02) }}
    
    {\footnotesize \texttt{ \} }}
    }%
}%

The code highlighted in blue is the implementation of the reward formalization constraint, \Cref{sec:dense_reward_form}.

\newpage

\subsection{Benchmark Environments Task descriptions}
\subsubsection{Grasp and Lift}
Task: The agent mush reach the cube position. Then grasp and lift the cube. Consider the task solved when the cube is at 0.5 height. There is no failure condition.
\begin{figure}[h]
    \centering
    \includegraphics[width=0.5\linewidth]{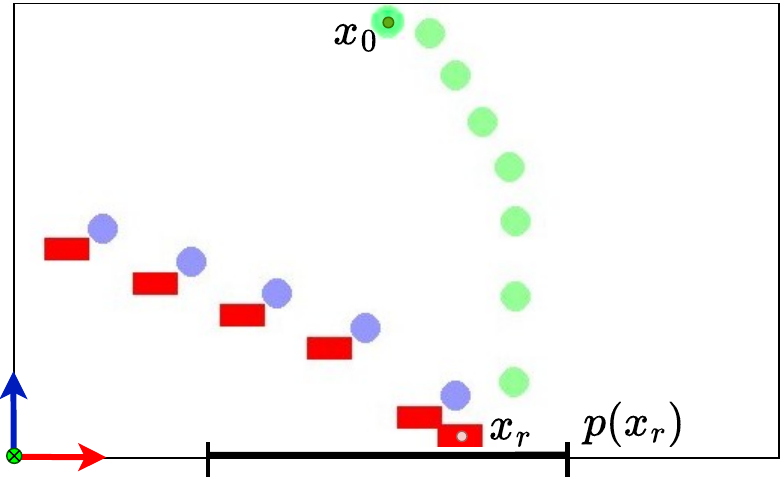}
    \label{fig:enter-label}
\end{figure}

\subsubsection{Grasp and Slide}
Task: The agent mush reach the cube position. Then grasp the cube. Finally, move the cube to the right (positive x direction). Consider the task solved when the cube x coordinate is $>$ 0.99. There is no failure condition.
\begin{figure}[h]
    \centering
    \includegraphics[width=0.5\linewidth]{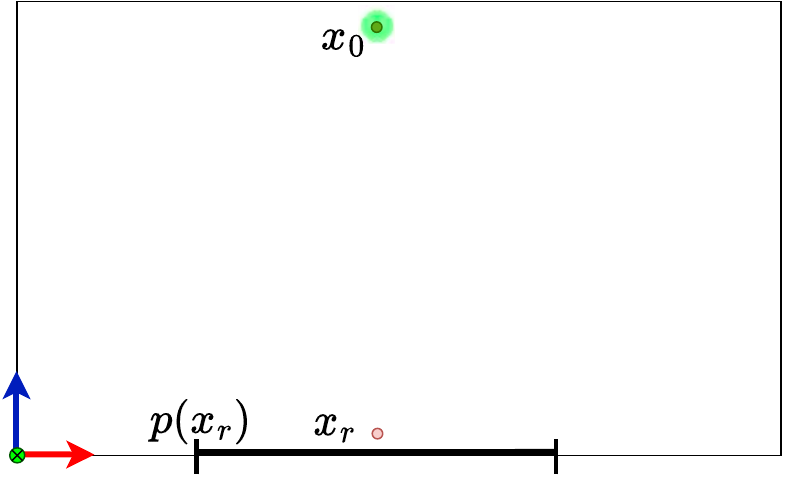}
    \label{fig:enter-label}
\end{figure}

\subsubsection{Place}
Task: The agent is holding the cube. Move the cube to (0, 0). Consider the task solved when the cube is at 0.05 distance from (0, 0) and is grasped. Consider the task failed when the agent doesn't grasp the cube.
\begin{figure}[h]
    \centering
    \includegraphics[width=0.5\linewidth]{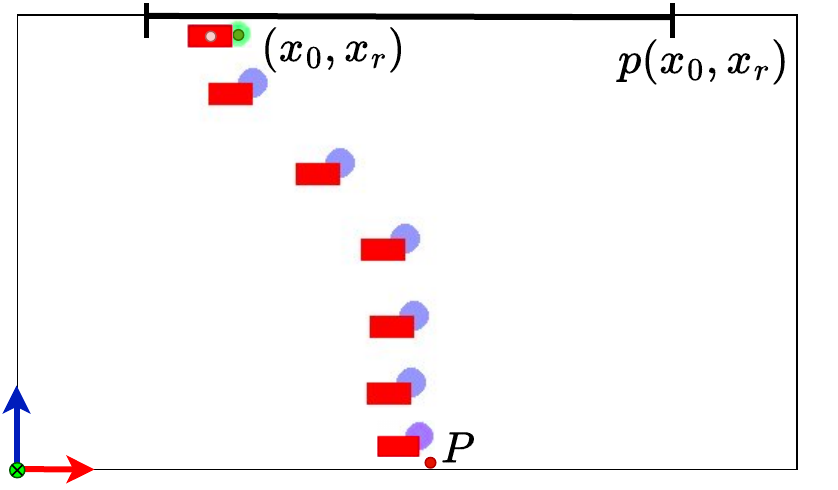}
    \label{fig:enter-label}
\end{figure}

\newpage

\subsubsection{Push Cube}
Task: The agent and the cube are on a narrow table. The agent mush reach the cube position. Then the agent must push the cube towards the positive x direction, without dropping the cube from the table. Consider the task solved when the cube is at x position $>$ 0.5. Consider the task failed when the cube is at z height $<$ table\_height.
\begin{figure}[h]
    \centering
    \includegraphics[width=0.5\linewidth]{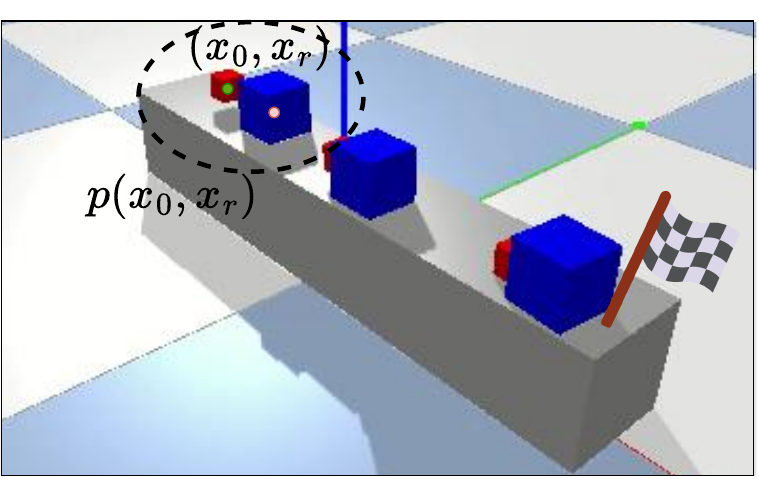}
    \label{fig:enter-label}
\end{figure}

\newpage
\subsection{Example of generated reward "fixed" by the proposed formalization}
As we argue in the paper, the weight imbalance issue is a major issue for reward generation with LLMs such as GPT-4.
In the included code, we have multiple examples of this being \emph{``fixed``} by the terminal reward defined in this formulation.
In the following, we consider one example of a reward function generated by ARCHIE with GPT-4, for the cube push task (\Cref{sec:bench_experiments}):
    \begin{equation}
        r(s_t, a_t) = -d + 10 \cdot contact + x_t,
    \end{equation}
    where $d$ is the distance between the two cubes, $contact$ is a boolean variable which is true when the cubes are in contact (part of the observation), and $x_t$ is the x coordinate of the blue cube. This is a typical example of reward with correct structure and unbalanced weights, since $d$ and $x_t$ are both in the order of $10^0$, while the weight of $contact$ is one order of magnitude greater.
    It is similar to the case in \Cref{fig:pushing_reward} of the manuscript.
    With the terminal term introduced in the novel formalization, the reward then becomes:
    \begin{equation}
        r(s_t, a_t, s_{t+1}) = -d + 10 \cdot contact + x_t + \underbrace{R_F(s_t, a_t) \Phi(s_{t+1})}_{\text{terminal term}},
        \label{eq:reward_example}
    \end{equation}
    with
    \begin{equation}
        \Phi(s_{t}) = \begin{cases}
            1 \quad \text{if } x_t > table\_lenght\\
            0 \quad \text{otherwise}.
        \end{cases}
    \end{equation}
    and $R_F(s_t, a_t)$ in the order of $10^2$ when $\Phi(s_{t+1})=1$.

    This example of reward was generated by GPT-4 in this Python code:

\noindent\fbox{%
    \parbox{\textwidth}{%
        {\footnotesize \texttt{def \_get\_info(self):}}\\
        {\footnotesize \texttt{~~~~cube\_pos, \_ = p.getBasePositionAndOrientation(self.cube2\_id)}}\\
        {\footnotesize \texttt{~~~~task\_solved = cube\_pos[0] > self.table\_lenght and cube\_pos[2] >= self.table\_height}}\\
        {\footnotesize \texttt{~~~~return \{'task\_solved': task\_solved\}}}\\  
        }%
}%

\noindent\fbox{%
    \parbox{\textwidth}{%
        {\footnotesize \texttt{def reward\_fun(self, observation, action):}}\\
        {\footnotesize \texttt{~~~~agent\_pos, \_ = p.getBasePositionAndOrientation(self.agent)}}\\ 
        {\footnotesize \texttt{~~~~cube\_pos, \_ = p.getBasePositionAndOrientation(self.cube2\_id)}}\\
        {\footnotesize \texttt{~~~~rewards\_dict = \{}}\\
        {\footnotesize \texttt{~~~~\# encourage proximity}}\\
        {\footnotesize \texttt{~~~~~~~~"distance\_to\_cube\_reward": -np.linalg.norm(np.array(cube\_pos[:3]) - np.array(agent\_pos[:3])), }}\\
        {\footnotesize \texttt{~~~~~~~~"contact\_reward": 10 * self.check\_contact(),  \# reward contact with the cube}}\\
        {\footnotesize \texttt{~~~~~~~~"x\_direction\_push\_reward": cube\_pos[0],  \# reward pushing in the x-direction}}\\
        {\footnotesize \texttt{~~~~\}}}\\
        {\footnotesize \texttt{~~~~total\_bonuses = sum([val if val > 0 else 0 for val in rewards\_dict.values()])}}\\
        {\footnotesize \texttt{~~~~total\_bonuses = max(total\_bonuses, 1)}}\\
        {\footnotesize \texttt{~~~~info = self.\_get\_info()}}\\
        {\footnotesize \texttt{~~~~total\_shaping = sum(rewards\_dict.values())}}\\
        {\footnotesize \texttt{~~~~task\_solved\_reward = 10 * self.\_max\_episode\_steps * total\_bonuses * info['task\_solved']}}\\
        {\footnotesize \texttt{~~~~reward = total\_shaping + task\_solved\_reward}}\\
        {\footnotesize \texttt{~~~~rewards\_dict['task\_solved\_reward'] = task\_solved\_reward}}\\
        {\footnotesize \texttt{~~~~return reward, rewards\_dict}}\\       
        }%
}%
\\
In \Cref{sec:bench_experiments}, \Cref{fig:pushing_reward} shows the learning curves of SAC agents trained with this reward and with its version without the terminal term.
Note Agents trained with the terminal reward term have episode termination enabled when the task is solved. Instead, the agents trained without the terminal term, i.e. the reward completely generated by GPT-4, do not have episode termination on completion enabled

\newpage
\newpage

\newpage
\subsection{Reward weights unbalance example}
\begin{figure}[h]
    \centering
    \subfloat{\includegraphics[width=0.5\columnwidth]{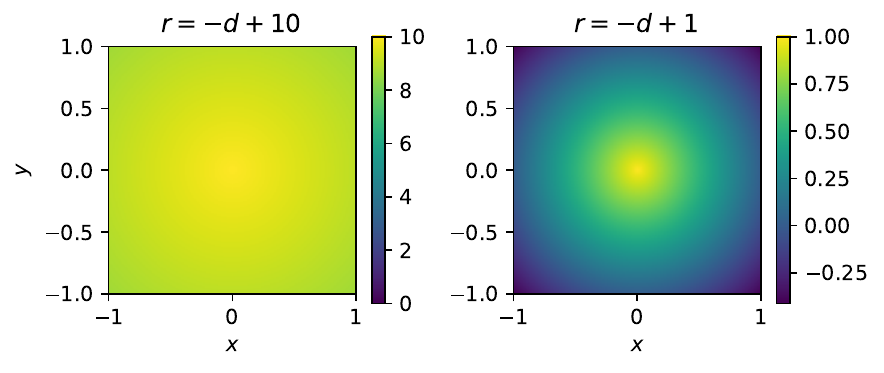}}%
    
    \subfloat{\includegraphics[width=0.5\columnwidth]{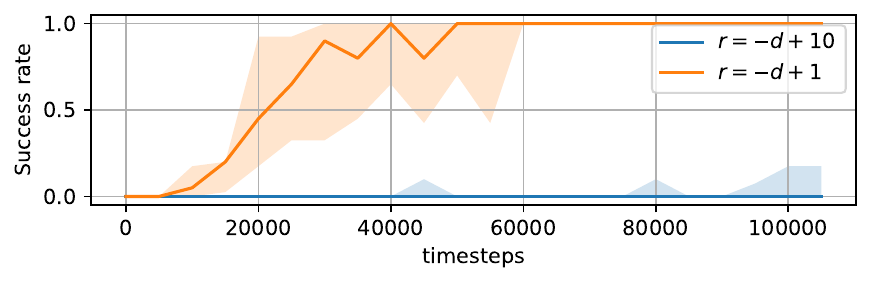}}
    \caption{Top: Rewards landscapes for two example rewards. Bottom: statistics of success rates of the SAC agents trained with the two rewards. }
    \label{fig:pushing_reward}
\end{figure}

Referring to the reward generation with LLMs described in \Cref{sec:nl2r}, with the available context, the shaping terms generated by the LLM are generally coherent with the task requirements. Nevertheless, there is no guarantee that the different generated terms are numerically coherent with each other or balanced, as discussed in other works. 
As an example, for a pushing task, where the agent is required to move an object from a starting position to a target position, one possible reward function that LLMs are capable of generating is:
\begin{equation}
    r(s_t, a_t) = - d + \begin{cases}
        b \text{ \hspace{0.2cm} if the agent is touching the object}\\
        0 \text{  \hspace{0.2cm} otherwise}
    \end{cases}
    \label{eq:pushing_reward}
\end{equation}
where $d$ is the distance between the object and the target position, while $b \in \mathbb{N^+}$ is a bonus for touching the object. \\
The bonus term $b$ can help guide exploration, but even simple rewards require tuning. If $b$ is too large relative to $d$, the reward landscape becomes too flat, leading to suboptimal learning. We illustrate this with a 2D example where a point-agent moves toward the origin. Using the reward from \cref{eq:pushing_reward}, we compare SAC training with $b=10$ and $b=1$. As shown in \Cref{fig:pushing_reward}, a large $b$ flattens the reward landscape, hindering learning, while a smaller $b$ offers clearer guidance and better performance. This highlights that even well-designed rewards can fail without proper weighting.

\newpage
\subsection{Robotic manipulation task descriptions and generated rewards}

\subsubsection{CubeSlide}
\emph{The robot gripper is close to the blue cube, touch it with the gripper fingers and push it close to the red cube. Consider the task solved if the distance between the cubes is less than 0.04 meters. Consider the task failed if the distance of the end effector from the blue cube is more than 0.1m.}
\begin{lstlisting}[language=Python]
def reward_fun(self, observation, action):
    rewards_dict = {}
    end_effector_left_pose = np.array(sim_utils.getLinkPose(self.robot_id, self.end_effector_left_idx))
    blue_cube_pose = np.array(sim_utils.getObjPose(self.objects_id['blue_cube']))
    red_cube_pose = np.array(sim_utils.getObjPose(self.objects_id['red_cube']))
    
    dist_blue = np.linalg.norm(np.array(blue_cube_pose[:3]) - np.array(end_effector_left_pose[:3]))
    dist_red_blue = np.linalg.norm(np.array(red_cube_pose[:3]) - np.array(blue_cube_pose[:3]))
    
    finger_1, finger_2, total_contact_force = self.check_fingers_touching(self.objects_id['blue_cube'])
    rewards_dict["dist_blue"] = -dist_blue
    rewards_dict["dist_red_blue"] = -dist_red_blue
    rewards_dict["finger_contacts_reward"] = 1 * finger_1 + 1 * finger_2
    
    info = self._get_info()   
    total_bonuses = sum([rewards_dict[k] if rewards_dict[k] > 0 else 0 for k in rewards_dict.keys()])
    total_bonuses = max(total_bonuses, 1)
    total_shaping = sum([rewards_dict[k] for k in rewards_dict.keys()])
    task_solved_reward = 10 * self._max_episode_steps * total_bonuses * info['task_solved']
    reward = total_shaping + task_solved_reward
    rewards_dict['task_solved_reward'] = task_solved_reward

    return reward, rewards_dict

def _get_info(self):
    blue_cube_pose = np.array(sim_utils.getObjPose(self.objects_id['blue_cube']))
    red_cube_pose = np.array(sim_utils.getObjPose(self.objects_id['red_cube']))
    dist_red_blue = np.linalg.norm(np.array(red_cube_pose[:3]) - np.array(blue_cube_pose[:3]))
    
    task_solved = dist_red_blue < 0.04
    return {'task_solved': task_solved}
  
def termination_condition(self):
    end_effector_left_pose = np.array(sim_utils.getLinkPose(self.robot_id, self.end_effector_left_idx))
    blue_cube_pose = np.array(sim_utils.getObjPose(self.objects_id['blue_cube']))
    dist_blue = np.linalg.norm(np.array(blue_cube_pose[:3]) - np.array(end_effector_left_pose[:3]))
    
    return dist_blue > 0.1
\end{lstlisting}

\newpage

\subsubsection{CubePick}
\emph{The robot gripper is close to a blue cube and a red cube. Grasp the blue cube with both fingers, then move the blue cube upwards of 0.05 meters. Consider the task completed when the blue cube is lifted by 0.05m over its loading height. Consider the task failed if the distance of the end effector from the blue cube is more than 0.1m or the red cube is moved from its loading position of 0.005m or more.}
\begin{lstlisting}[language=Python]
def reward_fun(self, observation, action):
    rewards_dict = {}  # Defines the reward components

    # getting the poses both the cubes and left end effector
    blue_cube_pose = np.array(sim_utils.getObjPose(self.objects_id['blue_cube']))
    red_cube_pose = np.array(sim_utils.getObjPose(self.objects_id['red_cube']))
    end_effector_left_pose = np.array(sim_utils.getLinkPose(self.robot_id, self.end_effector_left_idx))

    # calculating distances
    distance_blue = np.linalg.norm(np.array(blue_cube_pose[:3]) - np.array(end_effector_left_pose[:3]))
    distance_red = np.linalg.norm(np.array(red_cube_pose[:3]) - self.objects_dict['red_cube']['load_pos'])

    # calculating height difference of the blue cube from the initial position
    height_difference_blue = blue_cube_pose[2] - self.objects_dict['blue_cube']['load_pos'][2]

    # finger contact with blue cube
    finger_1_blue, finger_2_blue, _ = self.check_fingers_touching(self.objects_id['blue_cube'])

    rewards_dict = {
        "negative_distance_blue_cube_reward": - distance_blue,  # encourage getting close to the blue cube
        "negative_distance_red_cube_penalty": - distance_red if distance_red > 0.005 else 0,
        # negative reward if red cube is moved
        "finger_contacts_reward": 3 * (finger_1_blue + finger_2_blue),  # encourage finger contact with blue cube
        "blue_cube_height_reward": height_difference_blue if blue_cube_pose[2] >
                                                             self.objects_dict['blue_cube']['load_pos'][2] else 0,
        # reward if blue cube is lifted
    }

    total_bonuses = sum([rewards_dict[k] if rewards_dict[k] > 0 else 0 for k in rewards_dict.keys()])
    total_bonuses = max(total_bonuses, 1)
    info = self._get_info()
    total_shaping = sum([rewards_dict[k] for k in rewards_dict.keys()])
    task_solved_reward = 10 * self._max_episode_steps * total_bonuses * info['task_solved']

    reward = total_shaping + task_solved_reward

    rewards_dict['task_solved_reward'] = task_solved_reward

    return reward, rewards_dict

def _get_info(self):
    blue_cube_pose = np.array(sim_utils.getObjPose(self.objects_id['blue_cube']))
    red_cube_pose = np.array(sim_utils.getObjPose(self.objects_id['red_cube']))
    is_task_solved = blue_cube_pose[2] > self.objects_dict['blue_cube']['load_pos'][
        2] + 0.05  # task is solved if the blue cube is lifted over 0.05m
    return {'task_solved': is_task_solved}

def termination_condition(self):
    blue_cube_pose = np.array(sim_utils.getObjPose(self.objects_id['blue_cube']))
    red_cube_pose = np.array(sim_utils.getObjPose(self.objects_id['red_cube']))
    end_effector_left_pose = np.array(sim_utils.getLinkPose(self.robot_id, self.end_effector_left_idx))
    distance_blue = np.linalg.norm(np.array(blue_cube_pose[:3]) - np.array(end_effector_left_pose[:3]))
    distance_red = np.linalg.norm(np.array(red_cube_pose[:3]) - self.objects_dict['red_cube']['load_pos'])

    # Conditions for termination: distance from blue cube > 0.1 or red cube is moved by more than 0.005m
    return distance_blue > 0.1 or distance_red > 0.005
\end{lstlisting}

\newpage

\subsubsection{CubeStack}
\emph{The robot gripper is holding a blue cube, and a red cube is placed on a surface very close to the gripper. Do not drop the clue cube, keep the gripper fingers in contact with the blue cube. Place the blue cube on top of the red cube. Consider the task completed when the distance between the two cubes in the x-y plane is less than 0.005m, the absolute difference between the two cubes height is less or equal 0.0255m, the red cube is within 0.005m from its loading position. Consider the task failed if the gripper looses contact with the blue cube or the red cube is moved from its loading position of 0.005m or more, or the two cubes are further than at loading time.}
\begin{lstlisting}[language=Python]
def reward_fun(self, observation, action):
    # Initialize reward and reward components dictionary
    reward = 0.0
    # Calculate the pose for the blue and red cubes
    blue_cube_pose = np.array(sim_utils.getObjPose(self.objects_id['blue_cube']))
    red_cube_pose = np.array(sim_utils.getObjPose(self.objects_id['red_cube']))
    # Get the position and orientation of the left end effector
    end_effector_left_pose = np.array(sim_utils.getLinkPose(self.robot_id, self.end_effector_left_idx))
    # Check if the robot's fingers are touching the blue cube
    finger_1, finger_2, total_contact_force = self.check_fingers_touching(self.objects_id['blue_cube'])
    # Compute distance between the two cubes in the x-y plane and the height difference
    xy_distance = np.linalg.norm(blue_cube_pose[:2] - red_cube_pose[:2])
    height_diff = np.abs(blue_cube_pose[2] - red_cube_pose[2])
    # Compute the distance of the red cube from its loading position
    distance_red_cube_from_load_pos = np.linalg.norm(red_cube_pose[:3] - self.objects_dict['red_cube']['load_pos'])
    # Define the reward components
    rewards_dict = {
        "distance_to_red_cube_reward": -10*xy_distance, 
        "height_diff_reward": -10*height_diff,
        "finger_contacts_reward": int(finger_1) + int(finger_2),
        "distance_red_cube_from_load_pos_penalty": -distance_red_cube_from_load_pos if distance_red_cube_from_load_pos > 0.005 else 0
    }
    # Calculate total bonuses and shaping
    total_bonuses = sum([rewards_dict[k] if rewards_dict[k] > 0 else 0 for k in rewards_dict.keys()])
    total_bonuses = max(total_bonuses, 1)
    total_shaping = sum([rewards_dict[k] for k in rewards_dict.keys()])
    # Access the information on whether the task has been solved
    info = self._get_info()
    task_solved_reward = 10 * self._max_episode_steps * total_bonuses * info['task_solved']
    # Calculate total reward
    reward = total_shaping + task_solved_reward
    rewards_dict['task_solved_reward'] = task_solved_reward 
    # Return reward and reward components dictionary
    return reward, rewards_dict

def _get_info(self):
    # Calculate the pose for the blue and red cubes
    blue_cube_pose = np.array(sim_utils.getObjPose(self.objects_id['blue_cube']))
    red_cube_pose = np.array(sim_utils.getObjPose(self.objects_id['red_cube']))

    # Compute distance between the two cubes in the x-y plane and the height difference
    xy_distance = np.linalg.norm(blue_cube_pose[:2] - red_cube_pose[:2])
    height_diff = np.abs(blue_cube_pose[2] - red_cube_pose[2])

    # Compute the distance of the red cube from its loading position
    distance_red_cube_from_load_pos = np.linalg.norm(red_cube_pose[:2] - self.objects_dict['red_cube']['load_pos'][:2])

    task_solved = xy_distance < 0.005 and height_diff <= 0.0255 and distance_red_cube_from_load_pos < 0.005
    return {'task_solved': task_solved}

def termination_condition(self):
    # Check if the robot's fingers are touching the blue cube
    finger_1, finger_2, _ = self.check_fingers_touching(self.objects_id['blue_cube'])

    blue_cube_pose = np.array(sim_utils.getObjPose(self.objects_id['blue_cube']))
    red_cube_pose = np.array(sim_utils.getObjPose(self.objects_id['red_cube']))

    ditst_init  = np.linalg.norm(np.array(self.objects_dict['red_cube']['load_pos']) - self.objects_dict['blue_cube']['load_pos'])
    dist = np.linalg.norm(blue_cube_pose[:3] - red_cube_pose[:3])

    # Consider termination if the fingers are not touching the blue cube
    return not (finger_1 and finger_2) or dist > ditst_init
\end{lstlisting}

\newpage

\subsubsection{CubesTowerSlide}
\emph{The robot gripper is holding a blue cube, which is stacked on top of a red cube. Let go of the blue cube, move the tower of cubes to the right (negative direction of the y axis) of 0.1 meters, by touching the red cube. Avoid touching the blue cube. Consider the task solved if both cubes are moved to the right of 0.1 meters from their loading position. Consider the task failed if the x-y distance between the cubes is more than 0.01 meters, or the distance between end effector and the red cube is more than 0.05 meters.}
\begin{lstlisting}[language=Python]
def reward_fun(self, observation, action):
    rewards_dict = {} 

    blue_cube_pos = np.array(sim_utils.getObjPose(self.objects_id["blue_cube"]))
    red_cube_pos = np.array(sim_utils.getObjPose(self.objects_id["red_cube"]))
    end_effector_left_pose = np.array(sim_utils.getLinkPose(self.robot_id, self.end_effector_left_idx))

    distance_blue_red = np.linalg.norm(blue_cube_pos[:2] - red_cube_pos[:2])
    distance_effector_red = np.linalg.norm(end_effector_left_pose[:2] - red_cube_pos[:2])

    rewards_dict["distance_blue_red_penalty"] = -1000 if distance_blue_red > 0.01 else 0
    rewards_dict["distance_effector_red_penalty"] = -1000 if distance_effector_red > 0.05 else 0
    rewards_dict["move_right_bonus"] = (
        1000 * (self.objects_dict["blue_cube"]["load_pos"][1] - blue_cube_pos[1]) +
        1000 * (self.objects_dict["red_cube"]["load_pos"][1] - red_cube_pos[1]))

    total_bonuses = sum([rewards_dict[k] if rewards_dict[k] > 0 else 0 for k in rewards_dict.keys()])
    total_bonuses = max(total_bonuses, 1)
    info = self._get_info()
    total_shaping = sum([rewards_dict[k] for k in rewards_dict.keys()])
    task_solved_reward = 10 * self._max_episode_steps * total_bonuses * info['task_solved']

    reward = total_shaping + task_solved_reward

    rewards_dict['task_solved_reward'] = task_solved_reward

    return reward, rewards_dict

def _get_info(self):
    blue_cube_pos = np.array(sim_utils.getObjPose(self.objects_id["blue_cube"]))
    red_cube_pos = np.array(sim_utils.getObjPose(self.objects_id["red_cube"]))

    distance_blue_red = np.linalg.norm(blue_cube_pos[:2] - red_cube_pos[:2])
    task_solved = (
        distance_blue_red <= 0.01 and
        self.objects_dict["blue_cube"]["load_pos"][1] - blue_cube_pos[1] > 0.1 and
        self.objects_dict["red_cube"]["load_pos"][1] - red_cube_pos[1] > 0.1)

    return {'task_solved': task_solved}

def termination_condition(self):
    blue_cube_pos = np.array(sim_utils.getObjPose(self.objects_id["blue_cube"]))
    red_cube_pos = np.array(sim_utils.getObjPose(self.objects_id["red_cube"]))
    end_effector_left_pose = np.array(sim_utils.getLinkPose(self.robot_id, self.end_effector_left_idx))

    distance_blue_red = np.linalg.norm(blue_cube_pos[:2] - red_cube_pos[:2])
    distance_effector_red = np.linalg.norm(end_effector_left_pose[:2] - red_cube_pos[:2])

    return distance_blue_red > 0.01 or distance_effector_red > 0.05
\end{lstlisting}

\newpage

\subsubsection{VialGrasp}
\emph{The robot has just inserted a vial in a vial carrier, the gripper is very close to the vial. Grasp the vial, and lift out of the vial carrier, avoid to touch the vial carrier itself. Consider the task completed when the vial is lifted of 0.1m above loading height. Consider the task failed if the fingers loose contact with the vial, or the vial's distance in the x-y plane from its loading position is more than 0.025m.}
\begin{lstlisting}[language=Python]
def reward_fun(self, observation, action):
    rewards_dict = {}

    vial_pose = np.array(sim_utils.getObjPose(self.objects_id['vial']))
    vial_carrier_pose = np.array(sim_utils.getObjPose(self.objects_id['vial_carrier']))
    end_effector_left_pose = np.array(sim_utils.getLinkPose(self.robot_id, self.end_effector_left_idx))

    distance_to_vial_in_xy = np.linalg.norm(vial_pose[:2] - end_effector_left_pose[:2])

    height_diff = vial_pose[2] - self.objects_dict['vial']['load_pos'][2]

    finger_1, finger_2, total_contact_force = self.check_fingers_touching(self.objects_id['vial'])

    vial_not_moved = float(
        np.linalg.norm(vial_pose[:2] - np.array(self.objects_dict['vial']['load_pos'][:2])) < 0.025)

    rewards_dict = {
        "vial_height_reward": height_diff if height_diff > 0 else 0,
        "finger_contacts_reward": 1 * finger_1 + 1 * finger_2,  # encourage grasping the vial
        "distance_to_vial_penalty": -distance_to_vial_in_xy,  # encourage getting close to the vial
        "not_moved_reward": 10 * vial_not_moved
    }

    total_bonuses = sum([rewards_dict[k] if rewards_dict[k] > 0 else 0 for k in rewards_dict.keys()])
    total_bonuses = max(total_bonuses, 1)
    info = self._get_info()
    total_shaping = sum([rewards_dict[k] for k in rewards_dict.keys()])
    task_solved_reward = 10 * self._max_episode_steps * total_bonuses * info['task_solved']

    reward = total_shaping + task_solved_reward

    rewards_dict['task_solved_reward'] = task_solved_reward

    return reward, rewards_dict

def _get_info(self):
    vial_pose = sim_utils.getObjPose(self.objects_id['vial'])
    task_solved = vial_pose[2] > self.objects_dict['vial']['load_pos'][2] + 0.1
    return {'task_solved': task_solved}

def termination_condition(self):
    vial_pose = np.array(sim_utils.getObjPose(self.objects_id['vial']))
    end_effector_left_pose = np.array(sim_utils.getLinkPose(self.robot_id, self.end_effector_left_idx))

    finger_1, finger_2, _ = self.check_fingers_touching(self.objects_id['vial'])

    vial_moved = np.linalg.norm(vial_pose[:2] - np.array(self.objects_dict['vial']['load_pos'][:2])) > 0.025

    return not (finger_1 or finger_2) or vial_moved
\end{lstlisting}

\newpage

\subsubsection{CentrifugeInsertion}
\emph{The robot is holding a vial with its gripper. Lower the vial and insert it into the lab centrifuge, which is just below the vial. Consider the task solved when the vial is still touched by both gripper fingers, and the vial is at 0.08m height or below. Consider the task failed if the fingers loose contact with the vial, or the vial's distance in the x-y plane from its loading position is more than 0.04m.}
\begin{lstlisting}[language=Python]
def reward_fun(self, observation, action):
    vial_id = self.objects_id['vial']
    vial_pose = np.array(sim_utils.getObjPose(vial_id))
    centrifuge_id = self.objects_id['centrifuge']
    distance = np.linalg.norm(np.array(vial_pose[:2]) - np.array(self.objects_dict['vial']['load_pos'][:2]))

    finger_1, finger_2, total_contact_force = self.check_fingers_touching(vial_id)

    rewards_dict = {
        "finger_contacts_reward":  1 * finger_1 + 1 * finger_2,
        "vial_height_penalty":  -10 * vial_pose[2] ,
        "vial_xy_movement_penalty": -10 * distance
    }

    total_bonuses = sum([rewards_dict[k] if rewards_dict[k] > 0 else 0 for k in rewards_dict.keys()])
    total_bonuses = max(total_bonuses, 1)
    info = self._get_info()
    total_shaping = sum([rewards_dict[k] for k in rewards_dict.keys()])
    task_solved_reward = 10 * self._max_episode_steps * total_bonuses * info['task_solved']

    reward = total_shaping + task_solved_reward

    rewards_dict['task_solved_reward'] = task_solved_reward

    return reward, rewards_dict

def _get_info(self):
    vial_id = self.objects_id['vial']
    vial_pose = np.array(sim_utils.getObjPose(vial_id))

    success_condition = vial_pose[2] <= 0.08
    failure_condition = not self.termination_condition()

    return {'task_solved': success_condition and not failure_condition}

def termination_condition(self):
    vial_id = self.objects_id['vial']
    vial_pose = np.array(sim_utils.getObjPose(vial_id))
    initial_vial_pos = np.array(self.objects_dict['vial']['load_pos'])

    fail_cond_xy_distance = np.linalg.norm(np.array(vial_pose[:2]) - np.array(initial_vial_pos[:2])) > 0.04
    fail_cond_touching = not self.check_fingers_touching(vial_id)[0] or not self.check_fingers_touching(vial_id)[1]

    return fail_cond_xy_distance or fail_cond_touching
\end{lstlisting}

\newpage

\subsubsection{BimanualPenInTape}
\emph{The robot is holding a pen with its left gripper, and a tape with its right gripper. The tape has a hole in the middle. Keep the tape grasped by the right gripper, do not loose contact between the right gripper fingers and the cup. The same for the pen and the left gripper. Insert the pen into the hole of the tape. Consider the task solved when both objects are grasped, and the pen is at a distance from the tape of 0.005 or less. Consider the task failed if no finger of the right gripper is in contact with the tape, or no finger of the left gripper is in contact with the pen, or the distance between the objects in the horizontal plane is more than 0.025 meters.}
\begin{lstlisting}[language=Python]
def reward_fun(self, observation, action):
    # Define an empty dictionary for rewards
    rewards_dict = {}

    # Get pose of tape and pen
    tape_pose = np.array(sim_utils.getObjPose(self.objects_id["tape"]))
    pen_pose = np.array(sim_utils.getObjPose(self.objects_id["pen"]))

    # Get the distance between the pen and the tape
    distance = np.linalg.norm(tape_pose[:3] - pen_pose[:3])

    # Check the contacts of the fingers with tape and pen
    left_finger_1, left_finger_2, _, _ = self.check_fingers_touching(self.objects_id["pen"])
    _, _, right_finger_1, right_finger_2 = self.check_fingers_touching(self.objects_id["tape"])

    # Reward/punish different aspects of the task
    rewards_dict["distance_to_tape_reward"] = - 10 * distance
    rewards_dict["left_gripper_contacts_reward"] = 1 * left_finger_1 + 1 * left_finger_2
    rewards_dict["right_gripper_contacts_reward"] = 1 * right_finger_1 + 1 * right_finger_2

    # If distance is greater than 0.25 meters, give a negative reward
    rewards_dict["distance_penalty"] = -10 if distance > 0.25 else 0.0

    # Calculate total bonuses
    total_bonuses = sum([rewards_dict[k] if rewards_dict[k] > 0 else 0 for k in rewards_dict.keys()])
    total_bonuses = max(total_bonuses, 1)

    # Get task solved information and calculate total shaping
    info = self._get_info()
    total_shaping = sum([rewards_dict[k] for k in rewards_dict.keys()])
    
    # Define reward when task is solved
    task_solved_reward = 10 * self._max_episode_steps * total_bonuses * info['task_solved']
    # Sum total shaping and task solved reward
    reward = total_shaping + task_solved_reward
    # Add task solved reward to rewards dictionary
    rewards_dict['task_solved_reward'] = task_solved_reward
    return reward, rewards_dict

def _get_info(self):
    # Get the pose of the left and right arms
    tape_pose = np.array(sim_utils.getObjPose(self.objects_id["tape"]))
    pen_pose = np.array(sim_utils.getObjPose(self.objects_id["pen"]))
    # Check finger contacts with pen and tape
    left_finger_1, left_finger_2, _, _ = self.check_fingers_touching(self.objects_id["pen"])
    _, _, right_finger_1, right_finger_2 = self.check_fingers_touching(self.objects_id["tape"])
    # Find distance between pen and tape
    distance = np.linalg.norm(tape_pose[:3] - pen_pose[:3])
    # If all conditions are met, consider the task solved
    task_solved = (left_finger_1 and left_finger_2 and right_finger_1 and right_finger_2 and distance <= 0.005)
    return {"task_solved": task_solved}

def termination_condition(self):
    # Get pose of tape and pen
    tape_pose = np.array(sim_utils.getObjPose(self.objects_id["tape"]))
    pen_pose = np.array(sim_utils.getObjPose(self.objects_id["pen"]))
    # Check finger contacts with pen and tape
    left_finger_1, left_finger_2, _, _ = self.check_fingers_touching(self.objects_id["pen"])
    _, _, right_finger_1, right_finger_2 = self.check_fingers_touching(self.objects_id["tape"])
    # Calculate distance between pen and tape
    distance = np.linalg.norm(tape_pose[:2] - pen_pose[:2])
    # If any condition for failure is met, terminate
    if not (left_finger_1 and left_finger_2) or not (right_finger_1 and right_finger_2) or distance > 0.025:
        return True
    return False
\end{lstlisting}

\newpage

\subsubsection{CubeInCup}
\emph{The robot is holding a cube with its left gripper, and a cup with its right gripper. Keep the cup grasped by the right gripper, do not loose contact between the right gripper fingers and the cup. Place the cube into the cup. Consider the task solved when the cup is grasped by the gripper fingers of the right gripper, the cube is at a distance from the cup of 0.025 or less. Consider the task failed if no finger of the right gripper is in contact with the cup, or the distance between the left end effector and the cube is more than 0.2 meters.}
\begin{lstlisting}[language=Python]
def reward_fun(self, observation, action):
    rewards_dict = {}

    cube_id = self.objects_id["cube"]
    cup_id = self.objects_id["cup"]

    cube_pose = np.array(sim_utils.getObjPose(cube_id))
    cup_pose = np.array(sim_utils.getObjPose(cup_id))

    end_effector_left_pose = np.array(sim_utils.getLinkPose(self.robot_id, self.end_effector_left_idx))

    cup_cube_distance = np.linalg.norm(cube_pose[:3] - cup_pose[:3])
    left_effector_cube_distance = np.linalg.norm(cube_pose[:3] - end_effector_left_pose[:3])

    _, _, right_finger_1, right_finger_2 = self.check_fingers_touching(cup_id)

    rewards_dict = {
        "cube_cup_distance_reward": -cup_cube_distance,
        "cube_left_effector_distance_penalty": -left_effector_cube_distance,
        "gripper_contact_reward": 1 * right_finger_1 + 1 * right_finger_2
    }

    total_bonuses = sum([rewards_dict[k] if rewards_dict[k] > 0 else 0 for k in rewards_dict.keys()])
    total_bonuses = max(total_bonuses, 1)
    info = self._get_info()
    task_solved_reward = 10 * self._max_episode_steps * total_bonuses * info['task_solved']
    total_shaping = sum([rewards_dict[k] for k in rewards_dict.keys()])
    reward = total_shaping + task_solved_reward

    rewards_dict['task_solved_reward'] = task_solved_reward
    return reward, rewards_dict

def _get_info(self):
    cube_id = self.objects_id["cube"]
    cup_id = self.objects_id["cup"]

    cube_pose = np.array(sim_utils.getObjPose(cube_id))
    cup_pose = np.array(sim_utils.getObjPose(cup_id))

    cup_cube_distance = np.linalg.norm(cube_pose[:3] - cup_pose[:3])
    _, _, right_finger_1, right_finger_2 = self.check_fingers_touching(cup_id)

    task_solved = cup_cube_distance <= 0.025 and (right_finger_1 or right_finger_2)
    return {'task_solved': task_solved}

def termination_condition(self):
    cube_id = self.objects_id["cube"]
    cup_id = self.objects_id["cup"]
    end_effector_left_pose = np.array(sim_utils.getLinkPose(self.robot_id, self.end_effector_left_idx))
    cube_pose = np.array(sim_utils.getObjPose(cube_id))
    _, _, right_finger_1, right_finger_2 = self.check_fingers_touching(cup_id)

    left_effector_cube_distance = np.linalg.norm(cube_pose[:3] - end_effector_left_pose[:3])
    gripper_cup_contact = (right_finger_1 or right_finger_2)

    return left_effector_cube_distance > 0.2 or not gripper_cup_contact
\end{lstlisting}

\newpage

\subsubsection{BimanualBoxLift}
\emph{The robot has its left and right end effectors over a spaghetti box. Reach the spaghetti box with both grippers and grasp it using all fingers of both grippers. Consider the task partially solved when all four fingers are touching the box. Consider the task solved if the spaghetti box is lifted of 0.1m over its loading height, while all fingers are in contact with it. Consider the task failed when either end effector is further from the spaghetti box than 0.2 meters.}
\begin{lstlisting}[language=Python]
def reward_fun(self, observation, action):
    rewards_dict = {}  # Defines the reward components

    spaghetti_box_id = self.objects_id['spaghetti_box']
    spaghetti_box_pose = np.array(sim_utils.getObjPose(spaghetti_box_id))

    end_effector_left_pose = np.array(sim_utils.getLinkPose(self.robot_id, self.end_effector_left_idx))
    end_effector_right_pose = np.array(sim_utils.getLinkPose(self.robot_id, self.end_effector_right_idx))

    left_distance = np.linalg.norm(end_effector_left_pose[:3] - spaghetti_box_pose[:3])
    right_distance = np.linalg.norm(end_effector_right_pose[:3] - spaghetti_box_pose[:3])

    left_finger_1, left_finger_2, right_finger_1, right_finger_2 = self.check_fingers_touching(spaghetti_box_id)
    rewards_dict['left_distance_reward'] = -left_distance
    rewards_dict['right_distance_reward'] = -right_distance
    rewards_dict['fingers_contacts_reward'] = 1 * (left_finger_1 + left_finger_2 + right_finger_1 + right_finger_2)
    rewards_dict['grasp_reward'] = 10 * (left_finger_1 and left_finger_2 and right_finger_1 and right_finger_2)
    h = (spaghetti_box_pose[2] - self.objects_dict['spaghetti_box']['load_pos'][2])
    rewards_dict['spaghetti_box_height_reward'] = 100 * h if h > 0 else 0

    rewards_dict['orient_reward'] = - np.linalg.norm(spaghetti_box_pose[3:])

    total_bonuses = sum([rewards_dict[k] if rewards_dict[k] > 0 else 0 for k in rewards_dict.keys()])
    total_bonuses = max(total_bonuses, 1)
    info = self._get_info()
    task_solved_reward = 10 * self._max_episode_steps * total_bonuses * info['task_solved']
    total_shaping = sum([rewards_dict[k] for k in rewards_dict.keys()])

    reward = total_shaping + task_solved_reward

    rewards_dict['task_solved_reward'] = task_solved_reward

    return reward, rewards_dict

def _get_info(self):
    spaghetti_box_id = self.objects_id['spaghetti_box']
    spaghetti_box_pose = np.array(sim_utils.getObjPose(spaghetti_box_id))
    left_finger_1, left_finger_2, right_finger_1, right_finger_2 = self.check_fingers_touching(spaghetti_box_id)

    return {'task_solved': (spaghetti_box_pose[2] > self.objects_dict['spaghetti_box']['load_pos'][2] + 0.1 and
                            left_finger_1 and left_finger_2 and right_finger_1 and right_finger_2)}
                          # and np.linalg.norm(spaghetti_box_pose[3:]) < 0.5}

def termination_condition(self):
    spaghetti_box_id = self.objects_id['spaghetti_box']
    spaghetti_box_pose = np.array(sim_utils.getObjPose(spaghetti_box_id))

    end_effector_left_pose = np.array(sim_utils.getLinkPose(self.robot_id, self.end_effector_left_idx))
    end_effector_right_pose = np.array(sim_utils.getLinkPose(self.robot_id, self.end_effector_right_idx))

    left_distance = np.linalg.norm(end_effector_left_pose[:3] - spaghetti_box_pose[:3])
    right_distance = np.linalg.norm(end_effector_right_pose[:3] - spaghetti_box_pose[:3])

    return left_distance > 0.2 or right_distance > 0.2
\end{lstlisting}

\newpage

\subsubsection{BimanualHandover}
\emph{The robot is holding a box with its left gripper. Keep the box grasped by the left gripper, do not loose contact between the left gripper fingers and the box. Move the right gripper close to the box, grasp the box with the right gripper, and only then release the box with the left gripper. Consider the task solved when the box is grasped by both gripper fingers of the right gripper, and the left gripper's fingers have not been touching it for 50 time steps. Consider the task failed if no finger is in contact with the box, or the distance between the end effectors is more than 0.4 meters.}
\begin{lstlisting}[language=Python]
def reward_fun(self, observation, action):

    obj_pose = np.array(sim_utils.getObjPose(self.objects_id['box']))

    end_effector_left_pose = np.array(sim_utils.getLinkPose(self.robot_id, self.end_effector_left_idx))
    end_effector_right_pose = np.array(sim_utils.getLinkPose(self.robot_id, self.end_effector_right_idx))

    distance = np.linalg.norm(obj_pose[:3] - end_effector_right_pose[:3])

    left_finger_1, left_finger_2, right_finger_1, right_finger_2 = self.check_fingers_touching(self.objects_id['box'])

    rewards_dict = {
        "distance_to_box_reward": -distance, # Encourage getting close to the box
        "left_finger_touch_reward": 0.5 * (left_finger_1 + left_finger_2), # Reward for holding box
        "right_finger_touch_bonus": 2 * (right_finger_1 + right_finger_2), # Reward for grasping with right gripper
        "box_drop_penalty": -10 if not any([left_finger_1, left_finger_2, right_finger_1, right_finger_2]) else 0, # Penalty for dropping box
        "end_effectors_distance_penalty": -10 if distance > 0.4 else 0, # Penalty for large distance between end effectors.
    }

    total_bonuses = sum([rewards_dict[k] if rewards_dict[k] > 0 else 0 for k in rewards_dict.keys()])
    total_bonuses = max(total_bonuses, 1)
    info = self._get_info()
    total_shaping = sum([rewards_dict[k] for k in rewards_dict.keys()])
    task_solved_reward = 10 * self._max_episode_steps * total_bonuses * info['task_solved']

    reward = total_shaping + task_solved_reward

    rewards_dict['task_solved_reward'] = task_solved_reward

    return reward, rewards_dict

def _get_info(self):
    left_finger_1, left_finger_2, right_finger_1, right_finger_2 = self.check_fingers_touching(self.objects_id['box'])
    task_solved = right_finger_1 and right_finger_2 and not any([left_finger_1, left_finger_2]) and self.memory.get('count', 0) >= 50

    return {'task_solved': task_solved}

def termination_condition(self):
    left_finger_1, left_finger_2, right_finger_1, right_finger_2 = self.check_fingers_touching(self.objects_id['box'])
    dropped_box = not any([left_finger_1, left_finger_2, right_finger_1, right_finger_2])
    if dropped_box:
        return True
    if right_finger_1 and right_finger_2 and not any([left_finger_1, left_finger_2]):
        self.memory['count'] = self.memory.get('count', 0) + 1

    obj_pose = np.array(sim_utils.getObjPose(self.objects_id['box']))
    end_effector_right_pose = np.array(sim_utils.getLinkPose(self.robot_id, self.end_effector_right_idx))
    distance = np.linalg.norm(obj_pose[:3] - end_effector_right_pose[:3])
    if distance > 0.4:
        return True

    return False
\end{lstlisting}

\end{document}